\pdfoutput=1

\documentclass[11pt]{article}

\usepackage{acl}

\usepackage{times}
\usepackage{latexsym}

\usepackage[T1]{fontenc}

\usepackage[utf8]{inputenc}

\usepackage{microtype}

\usepackage{inconsolata}

\usepackage{graphicx}

\usepackage{lipsum}  
\usepackage{tabularx}
\usepackage{booktabs}

\usepackage{microtype}
\usepackage{type1cm}
\usepackage[american]{babel}
\usepackage{amssymb}
\usepackage{amsmath}
\usepackage{multirow}
\usepackage{xspace}
\DeclareMathAlphabet{\mathcalbf}{OMS}{pzc}{b}{n}
\usepackage{enumitem}
\usepackage{colortbl}
\usepackage{xcolor}

\RequirePackage{color}
\definecolor{darkgray}{gray}{0.40}
\definecolor{mediumgray}{gray}{0.60}
\definecolor{lightgray}{gray}{0.95}
\definecolor{ultralightgray}{gray}{0.98}
\definecolor{forestgreen}{rgb}{0.133, 0.545, 0.133}
\definecolor{orange}{rgb}{1, 0.86, 0.74}
\definecolor{lightergreen}{rgb}{0.95, 1, 0.88}
\definecolor{pastelblue}{rgb}{0.85, 0.92, 0.98}
\definecolor{pastelyellow}{rgb}{1.0, 0.95, 0.80}
\definecolor{pastelgray}{rgb}{0.94, 0.94, 0.96}

\usepackage{graphicx}
\DeclareGraphicsExtensions{.pdf,.ai,.jpg,.png}
\setkeys{Gin}{pagebox=artbox}
\graphicspath{{../acl26-edit-paper-figures/}}

\newcommand{\bsfigure}[3][]{%
	\begin{figure}[t]
		\centering
		\includegraphics[#1]{#2}
		\caption{#3}\label{#2}%
	\end{figure}
}

\newcommand{\hwfigure}[3][t!]{%
	\begin{figure*}[#1]
		\centering
		\includegraphics[scale=1.0]{#2}
		\caption{#3}\label{#2}%
	\end{figure*}
}

\RequirePackage{type1cm}
\RequirePackage{color}
\RequirePackage{soul}
\RequirePackage[normalem]{ulem}
\setstcolor{blue}
\definecolor{violet}{rgb}{0.5,0.0,0.5}

\newsavebox\bscombox
\newcommand{\bscom}[3][]{%
    \sbox{\bscombox}{\fontsize{8}{9}\selectfont#1#2#3}
    \noindent
    \st{#2}{\selectfont
        \color{blue}#3\ifx\\#1\\\else{\fontsize{8}{9}\selectfont\color{violet}[#1]}\fi
    }
}

\usepackage{tikz}

\begin{document}

\title{
Teaching LLMs Human-Like Editing of Inappropriate Argumentation \\ via Reinforcement Learning
}




\author{
	Timon Ziegenbein \\
	Leibniz University Hannover \\
	\texttt{t.ziegenbein@ai.uni-hannover.de} \And
	Maja Stahl \\
	Leibniz University Hannover \\
	\texttt{m.stahl@ai.uni-hannover.de} \AND
	Henning Wachsmuth \\
	Leibniz University Hannover, L3S Research Center \\
	\texttt{h.wachsmuth@ai.uni-hannover.de}
}




\date{}

\maketitle

\begin{abstract}

Editing human-written text has become a standard use case of large language models (LLMs), for example, to make one's arguments more appropriate for a discussion. Comparing human to LLM-generated edits, however, we observe a mismatch in editing strategies: While LLMs often perform multiple scattered edits and tend to change meaning notably, humans rather encapsulate dependent changes in self-contained, meaning-preserving edits. In this paper, we present a reinforcement learning approach that teaches LLMs human-like editing to improve the appropriateness of arguments. Our approach produces self-contained sentence-level edit suggestions that can be accepted or rejected independently. We train the approach using group relative policy optimization with a multi-component reward function that jointly optimizes edit-level semantic similarity, fluency, and pattern conformity as well as argument-level appropriateness. In automatic and human evaluation, it outperforms competitive baselines and the state of the art in human-like editing, with multi-round editing achieving appropriateness close to full rewriting.

\end{abstract}

\section{Introduction}
\label{sec:intro}

\bsfigure{human-likeness-example}{Two similar human appropriateness edits of two exemplary sentences along with their edit diffs. We argue that three criteria make edits human-like: (a) semantic \emph{similarity} to the original sentence, (b) \emph{fluency} in dependent edit operations (\emph{dep.}), and (c) \emph{conformity} to typical keep/delete/add/substitute edit patterns. Our reward model teaches LLMs to edit an argument's sentences accordingly while optimizing its appropriateness.}

Automated systems based on large language models (LLMs) are increasingly used in everyday life for text production and optimization tasks \cite{zhang-etal-2025-trending}. A typical use case is argumentative writing, which is omnipresent across education, public discourse, and online debate. Arguments are often presented in ways that violate norms of appropriateness --- through offensive language \citep{wulczyn-etal-2017-ex}, 
overly emotional appeals \citep{walton-2010-appeal}, unclear reasoning, or personal attacks \citep{habernal-etal-2018-argument}. Such inappropriateness not only harms an argument's persuasive effect, but it also impedes critical discussions. 

For general text optimization, multi-step methods that generate explicit edit suggestions exist \citep{du-etal-2022-understanding-iterative, raheja-etal-2023-coedit}, but they require substantial supervised training data with edit-level annotations, which is not available for inappropriate arguments.
Recently, \citet{ziegenbein-etal-2024-llm} explored how to align LLMs' output behavior to make inappropriate arguments appropriate. However, their approach generates full rewrites that optimize for argument-level metrics rather than individual edits or edit suggestions. While the performed edits can be regained from a diff algorithm, we observe that they often yield scattered, interdependent changes that cannot be applied selectively and notably affect the meaning of the original argument.
In contrast, humans tend to produce self-contained, meaning-preserving edit suggestions that encapsulate dependent changes in a single contiguous operation, as illustrated in Figure~\ref{human-likeness-example}: Both edits remain similar to the original sentence while ensuring fluency and following comparable edit patterns.

This paper presents an approach that teaches LLMs to generate human-like edit suggestions for inappropriate arguments. Via reinforcement learning, the approach directly produces self-contained and meaning-preserving edit suggestions. Drawing inspiration from text style transfer \citep{jin-etal-2022-deep}, where systems typically optimize document-level metrics for semantic similarity, fluency, and style \citep{luo-etal-2019-dual}, we explicitly optimize these criteria through reward modeling at the edit level --- unlike any existing method. We further extend the traditional style transfer objectives by introducing an edit pattern conformity metric that evaluates whether an LLM's delete, add, and substitute operation sequences match human editing behavior.

We train our policy using group relative policy optimization \citep{shao-etal-2024-deepseek}, instantiated in a Llama-3.1-8B-Instruct model \citep{grattafiori-etal-2024-llama}. Our reward function trades three edit-level quality metrics (similarity, fluency, and conformity) evaluated at the edit level against argument-level appropriateness improvement. We conducted automatic and human evaluation on arguments annotated for inappropriateness \citep{ziegenbein-etal-2023-modeling}, finding that our approach outperforms competitive baselines, including the state-of-the-art rewriting approach of \citet{ziegenbein-etal-2024-llm}. It generates substantially more human-like edit suggestions, while multi-round iterative editing closes in on the argument-level appropriateness of the best non-human-like rewriting approach.

Altogether, our contributions are threefold:%
\footnote{Code and data are available at \url{https://github.com/timonziegenbein/inappropriateness-editing}.}
\begin{itemize}
\setlength{\itemsep}{0pt}
    \item A reward model for edit-level quality criteria (semantic similarity, fluency, pattern conformity), explicitly aiming at human-likeness.
    \item The first reinforcement learning approach to generating human-like edit suggestions for inappropriate arguments.
    \item Empirical evidence that optimizing for edit-level quality substantially improves human-like edit generation while maintaining appropriateness improvement.
\end{itemize}

\section{Related Work}
\label{sec:related_work}

Text editing is a long-standing NLP task. Early work focused on cognitive models of writing and revision \citep{flower-etal-1980-dynamics, collins-etal-1980-framework, vaughan-etal-1986-model}, whereas modern approaches often frame text editing as a sequence-to-sequence task. 

To model the editing process, several works introduce taggers that predict edit operation sequences (add, delete, keep) and then operationalize them for tasks like grammatical error correction and text simplification \citep{malmi-etal-2019-lasertagger, mallinson-etal-2020-felix, omelianchuk-etal-2020-gector, stahlberg-etal-2020-seq2edits, dong-etal-2019-editnts}.
More recent work focuses on understanding the human revision process itself. \citet{yang-etal-2017-identifying} classify semantic edit intentions in Wikipedia. \citet{jiang-etal-2022-arxivedits} and \citet{du-etal-2022-understanding-iterative} present corpora of iterative revisions to better model the writing process. Closer to the task at hand, approaches exist that rewrite toxic content  \citep{nogueira-etal-2018-fighting, laugier-etal-2021-civil, he-etal-2023-prompt}. Building on instruction fine-tuning approaches such as Alpaca \citep{taori-etal-2023-alpaca} and Self-Instruct \citep{wang-etal-2023-selfinstruct}, \citet{raheja-etal-2023-coedit} and its multilingual successor \citet{raheja-etal-2024-medit} demonstrated that fine-tuning on diverse task-specific instructions yields state-of-the-art text editing systems. \citet{shu-etal-2023-rewritelm} and \citet{zeng-etal-2025-fineedit} also leverage instruction tuning for text rewriting and editing.
However, unlike all these supervised approaches that learn human-like editing only implicitly from parallel data, our work explicitly optimizes for human-like edits without requiring parallel data, making it applicable to a much wider range of edit-based revision tasks.

The revision dimension \emph{appropriateness} that we focus on is part of a widely-used taxonomy of argument quality \citep{wachsmuth-etal-2017-computational}. \citet{wachsmuth-werner-2020-intrinsic} study the computational assessment of all taxonomy dimensions, including appropriateness, whereas \citet{habernal-etal-2018-argument} and \citet{salminen-etal-2018-anatomy} address related quality issues in online discussions, namely fallacies and hate speech respectively.
More recently, \citet{ziegenbein-etal-2023-modeling} assess appropriateness specifically and introduce a fine-grained taxonomy of inappropriateness in argumentation as part of this, which we start from in our work.

While most argument quality research focuses on assessment, some work explores the generation of better arguments: \citet{skitalinskaya-etal-2022-claim} propose a generate-and-rank approach for claim optimization, and \citet{huber-etal-2025-clear} systematically evaluate argument rewriting with LLMs. \citet{stahl-etal-2025-arginstruct} train an instruction fine-tuned LLM specialized for argumentation tasks, including appropriateness assessment. Closest to our work, \citet{ziegenbein-etal-2024-llm, ziegenbein-etal-2024-objective} use reinforcement learning (RL) to rewrite inappropriate arguments. Unlike their approach, we teach LLMs \emph{human-like} editing via reinforcement learning.

Since RL generalizes well across tasks without explicit training data, aligning it with human preferences has become a popular technique for improving LLMs.
Starting with the works of \citet{christiano-etal-2017-deep} and \citet{ziegler-etal-2019-finetuning}, RL from human feedback (RLHF) has been used to improve language models on a variety of tasks, including summarization \citep{stiennon-etal-2020-learning}.
\citet{ouyang-etal-2022-training} show that RLHF can make models better at following instructions, using
proximal policy optimization (PPO) \citep{schulman-etal-2017-proximal}, a commonly used algorithm, also employed by \citet{ziegenbein-etal-2024-objective}.
Our approach uses the PPO-variant group-relative policy optimization (GRPO) \citep{shao-etal-2024-deepseek}.
While RL has been used for non-parallel style transfer \citep{xu-etal-2018-unpaired, gong-etal-2019-reinforcement, wu-etal-2019-hierarchical, luo-etal-2019-dual}, subjective bias correction \citep{madanagopal-etal-2023-reinforced}, and detoxification \citep{laugier-etal-2021-civil, logacheva-etal-2022-paradetox}, these approaches typically evaluate quality using document-level metrics for fluency, semantic similarity, and style. To our knowledge, we are the first to map fluency and semantic similarity to the edit level and to extend these metrics by studying the conformity of edit operation sequences. This enables us to use RL for the generation of human-like edits.

\section{Approach}
\label{sec:approach}

\hwfigure{approach}{Our reinforcement learning approach to human-like appropriateness editing, including the policy $\pi(E|a)$ for editing an argument $a$, the reward function $R$ based on the classifiers $c_{sim}$, $c_{flu}$, and $c_{con}$, and their interaction.}

This section presents our approach to generating human-like edit suggestions for inappropriate arguments based on reinforcement learning (RL). We use a large language model (LLM) as a policy that is trained with group relative policy optimization (GRPO) \citep{shao-etal-2024-deepseek} to generate a set of human-like edits for a given argument. An overview is given in Figure~\ref{approach}. We detail the approach in the following, whereas technical details of the employed reward classifiers and the GRPO training process can be found in Section~\ref{sec:reward-classifiers}.

\subsection{Problem Definition}

Given an argument $a$, our goal is to generate a set of edit suggestions $E = \{e_1, e_2, ..., e_k\}$. Each edit $e_i := (s_i, t_i)$ consists of a text span $s_i$ to rewrite and its replacement $t_i$. The goal is to find a set $E^*$ such that all its edit suggestions $e^*_i$ are \emph{human-like} and applying them to $a$ results in a new argument $a'$ that is more \emph{appropriate} than $a$. 

\paragraph{Human-Like Edits}
An edit suggestion $e := (s, t)$ is human-like if applying it to argument $a$ results in text that is \emph{semantically similar} to $a$, does not degrade \emph{fluency}, and is \emph{conform} in its deletions, additions, and substitutions to typical human edit patterns. Semantic similarity and fluency follow common metrics in style transfer tasks \citep{jin-etal-2022-deep}, but are adapted here to the edit level. Conformity refers to the edits' surface-level form, motivated by observing LLM-generated edits often exhibit different patterns than those of human editors \citep{ziegenbein-etal-2024-llm}, like keeping large spans of text with only a single token addition, deletion, or replacement (see Appendix~\ref{app:category-violations} for examples).

\paragraph{(In)Appropriateness}
Following \citet{ziegenbein-etal-2023-modeling}, we deem an argument inappropriate (in light of its discussion context) if it is missing commitment of its author to the discussion, uses toxic emotions, or is missing intelligibility.

\subsection{Reinforcement Learning}

Our RL policy, denoted as $\pi(E|a)$, is an LLM that generates a set of edit suggestions $E$ for a given argument $a$. To apply an edit $e_i$, we replace $s_i$ with $t_i$ in $a$ to produce $a'$. 
We use GRPO, a memory-efficient RL algorithm, to train our policy $\pi_\theta(E|a)$, which is parameterized by $\theta$. The objective of the training of GRPO is to find the parameters $\theta^*$ that maximize the expected reward:
\begin{eqnarray*}
\theta^* & := & \arg\max_\theta \mathbb{E}_{E \sim \pi_\theta(E|a)} [R(a, E)]
\end{eqnarray*}
Our overall reward function $R(a, E)$ combines several components evaluating aspects of the generated edit suggestions to make them \emph{human-like}.

\paragraph{Human-Likeness}

Let $a_i$ denote the sentence in argument $a$ that contains the edit span $s$. We operationalize the definition of human-like edits in a function $h(e, a_i)$ that returns 1 if edit $e$ is human-like when applied to sentence $a_i$, and 0 otherwise:
\begin{equation*}
h(e, a_i)  := \mathbb{I}(c_{sim}(e, a_i) \land c_{flu}(e, a_i) \land c_{con}(e, a_i))
\end{equation*}
where $\mathbb{I}$ is the truth interpretation function and $c_{sim}(e, a_i)$, $c_{flu}(e, a_i)$, and $c_{con}(e, a_i)$ are the binary outputs of three sentence-level classifiers---semantic similarity, fluency, and conformity---that evaluate whether the edit $e$ maintains these quality criteria when applied to $a_i$. These classifiers provide local quality signals that complement the global appropriateness reward. Details on their implementation are provided in Section~\ref{sec:reward-classifiers}.

\paragraph{Edit-Level Reward}

Given $h(e, a_i)$, we define the edit-level reward $r_{edit}(E, a)$ as the proportion of human-like edit suggestions in the set $E$:
\begin{eqnarray*}
r_{edit}(E, a) & := & \frac{1}{|E|} \sum_{e \in E} h(e, a_i)
\end{eqnarray*}
where $a_i$ is the sentence in $a$ containing $e$.

\paragraph{Argument-Level Reward}

The argument-level reward $r_{arg}(a, E)$ quantifies the appropriateness of the argument $a$ after applying all human-like edit suggestions. Let $E_{\text{HL}} \subseteq E$ be the set of human-like edit suggestions in $E$, and let $a'$ be the result of applying all edit suggestions in $E_{\text{HL}}$ to $a$. Then the argument-level reward is:
\begin{eqnarray*}
r_{arg}(a, E) & := & c_{app}(a') \cdot \mathbb{I}(|E_{\text{HL}}| > 0)
\end{eqnarray*}
where $c_{app}(a')$ is the output of an appropriateness classifier for the edited argument $a'$.

\paragraph{Overall Reward}

The overall reward of a set of edit suggestions $E$ is a weighted combination of the edit-level and argument-level rewards:
\begin{equation*}
R(a, E) := \alpha \cdot r_{arg}(a, E) + (1 - \alpha) \cdot r_{edit}(E, a)
\end{equation*}
where $\alpha$ controls the trade-off between argument-level appropriateness and edit-level quality. Note that this formulation provides both a direct signal from the edit-level rewards (even if the edits do not improve appropriateness) and an indirect signal through the argument-level reward, where we apply all edit suggestions that were classified as human-like by the edit-level reward classifiers.

\paragraph{Inference}

During inference, the trained policy $\pi_\theta$ generates edit suggestions fully autonomously. It can also be applied iteratively, feeding the revised argument $a'$ back as input for the next round and repeating until appropriateness converges. For ethical use, however, a human-in-the-loop deployment is recommended (see Section~\ref{sec:ethics}).

\section{Data}
\label{sec:data}

Our approach requires two types of datasets: (1) human revision data to train the edit-level classifiers $c_{sim}$, $c_{flu}$, and $c_{con}$ for human-likeness, and (2) appropriateness-annotated arguments to train and evaluate the GRPO model.

\subsection{Human Revisions}
\label{sec:classifier-data}

The IteraTeR dataset \citep{du-etal-2022-understanding-iterative} contains 172,692 human edits from iteratively revised text in Wikipedia, ArXiv, and Wikinews. Each edit is categorized as either meaning-changed or non-meaning-changed, with the latter assigned to \emph{Fluency}, \emph{Coherence}, \emph{Clarity}, \emph{Style}, or \emph{Other}. 

We base our semantic similarity, fluency, and conformity classifiers on the IteraTeR dataset. For the fluency classifier, we extract positive samples (edits that improve fluency) from IteraTeR's \emph{fluency} edits and generate negative samples (edits that degrade fluency) by reversing the edits, that is, by applying the inverse transformation to the improved text to recreate the original, less fluent version. This results in 7,804 instances, which we further augment with 9,519 instances from Gemini 2.5 Flash \citep{geminiteam2025geminifamilyhighlycapable} queried for fluency decisions during GRPO training using the prompt described in Appendix~\ref{sec:appendix-fluency-prompt}, resulting in 17,323 instances (6,559 positive and 10,764 negative examples).

\subsection{Appropriateness-Annotated Arguments}
\label{sec:argument-data}

We use the appropriateness corpus of \citet{ziegenbein-etal-2023-modeling}, extended by \citet{ziegenbein-etal-2024-llm}, which contains argumentative texts from online discussions annotated across 14 dimensions of appropriateness. The original corpus contains 1182 inappropriate and 1009 appropriate arguments. The extension adds 49,417 soft-labeled arguments from the Internet Argument Corpus v2 \citep{walker-etal-2012-corpus,abbott-etal-2016-internet} and the GAQCorpus \citep{ng-etal-2020-creating}, comprising 35,537 inappropriate and 13,880 appropriate arguments. The corpus distinguishes between appropriateness issues in four categories: \emph{toxic emotions} (deceptive or excessive emotional appeals), \emph{missing commitment} (lack of seriousness or openness), \emph{missing intelligibility} (unclear meaning or reasoning), and \emph{other reasons} (orthographic errors and additional issues).

We train the GRPO model on inappropriate arguments from the extended corpus and evaluate on inappropriate arguments from the original corpus.

\section{Experiments}
\label{sec:experimental-setup}

This section details our experimental setup to evaluate human-like argument appropriateness rewrites. We first introduce four reward classifiers, three operating on the sentence level: semantic similarity $c_{sim}(e, a_i)$, fluency $c_{flu}(e, a_i)$, and conformity $c_{con}(e, a_i)$, where $e$ is the edit and $a_i$ is the sentence containing the edit. The last, the appropriateness classifier $c_{app}(a')$, operates at the argument level. Then, we present the baselines against which we compare our approach, describe our training setup, and outline how we evaluate the approach.

\subsection{Reward Classifiers}
\label{sec:reward-classifiers}

The three edit-level classifiers (semantic similarity, fluency, and conformity) serve as minimal-requirement classifiers: an edit is considered human-like only if it passes all three requirements. Each classifier evaluates a necessary condition for edit quality, and all conditions must be satisfied for an edit to be accepted (i.e., human-like).

\paragraph{Semantic Similarity}

For $c_{sim}(e, a_i)$, we use Google's embedding model EmbeddingGemma-300M \citep{schechtervera-etal-2025-embeddinggemma} to measure the semantic similarity between the original sentence $a_i$ and the sentence after applying edit suggestion $e$. An edit is considered to preserve semantic similarity if the embedding similarity exceeds a threshold $\tau = .6757$, which corresponds to the 99th percentile of similarities computed on edits from the IteraTeR dataset. $\tau$ ensures that only edits with semantic similarity comparable to or better than 99\% of human edits are accepted.

\paragraph{Fluency}

For $c_{flu}(e, a_i)$, we train a binary classifier based on ModernBERT \citep{warner-etal-2025-smarter} to detect fluency degradation. The classifier takes as input the original sentence $a_i$ and the sentence after applying edit suggestion $e$ to determine whether the edit maintains or improves grammatical correctness. We train on the modified IteraTeR dataset and on Gemini augmentation (see Appendix~\ref{sec:appendix-fluency-prompt} for the prompt) as described in Section~\ref{sec:data}. 

Since fluency errors are detrimental to edits regardless of their quality in other dimensions, we optimize for high precision by selecting the model checkpoint that achieves the best precision on the validation data during training to ensure that accepted edits are grammatically correct. This addresses the interdependency problem discussed in Section~\ref{sec:intro}: by rejecting edit suggestions that would introduce grammatical errors when applied individually, $c_{flu}$ ensures that each edit can be applied independently without breaking text coherence. Our final classifier achieves a precision of 0.880 and an F$_1$-score of 0.835. Training details, result analysis, and a comparison with existing grammatical error correction approaches (rule-based, trained, and LLM prompting) adjusted to our task are provided in Appendix~\ref{sec:appendix-fluency}.

\paragraph{Pattern Conformity}

For $c_{con}(e, a_i)$, we quantify edit conformity using a language model trained on edit operation sequences. For each edit suggestion $e$ applied to sentence $a_i$, we tokenize the original sentence and the sentence after applying the edit, then compute the sequence of diff operations to transform the original into the edited version:
\begin{itemize}
\setlength{\itemsep}{-2pt}
	\item \textit{Keep} (tokens outside edit region),
	\item \textit{Keep-in-edit} (unchanged tokens within edit),
	\item \textit{Del} (deleted tokens),
	\item \textit{Add} (added tokens), and
	\item \textit{Substitute} (replaced tokens).
\end{itemize}

The training is semi-supervised---we leverage the inherent structure of edit sequences without explicit labels. We train a decoder-only transformer language model on all diff sequences from the IteraTeR dataset, predicting the next operation through cross-entropy loss, implicitly capturing surface-level token patterns and sequential conformity. During inference, we compute the perplexity of the edit operation sequence, where lower perplexity indicates greater conformity to edit patterns. An edit is considered conform if its perplexity is below the 99th percentile threshold from IteraTeR edits. Further training details are provided in Appendix~\ref{sec:appendix-conformity}.

\paragraph{Appropriateness}

For $c_{app}(a')$, we reuse the multilabel appropriateness classifier of \citet{ziegenbein-etal-2023-modeling} as our appropriateness reward model. This classifier operates at the argument level, evaluating the full argument consisting of one or multiple sentences. The classifier is trained to perform multilabel classification of 14 inappropriateness categories across three hierarchy levels. It outputs an inappropriateness score $s(a')$ for each category; we reverse this to obtain an appropriateness score for our reward model: $c_{app}(a') := 1 - s(a')$.

\subsection{Approaches and Baselines}
\label{sec:baselines}

We evaluate several methods for appropriateness rewriting, including our GRPO-based approach with edit-level classifiers as well as PPO-based baselines from \citet{ziegenbein-etal-2024-llm}. All approaches generate full argument rewrites; for comparison on edit-level metrics, we extract edit suggestions from the rewrites using \texttt{latexdiff}.%
\footnote{\url{https://ctan.org/pkg/latexdiff}}

\paragraph{Baselines}
We compare against the following PPO-based methods using Alpaca \citep{taori-etal-2023-alpaca}, following the original implementation of \citet{ziegenbein-etal-2024-llm}:
\begin{itemize}
\setlength{\itemsep}{0pt}
	\item \textit{PPO$_{\text{app}}$.} Alpaca optimized for appropriateness only using PPO, without semantic similarity constraints.
	\item \textit{PPO$_{\text{app$<$sim}}$.} PPO training with appropriateness weighted lower than semantic similarity.
	\item \textit{PPO$_{\text{app$=$sim}}$.} With equal weighting between appropriateness and semantic similarity.
	\item \textit{PPO$_{\text{app$>$sim}}$.} With appropriateness weighted higher than semantic similarity.
\end{itemize}

\paragraph{Approach}
Our approach uses Llama-3.1-8B-Instruct as the policy $\pi_\theta$ with GRPO training:
\begin{itemize}
\setlength{\itemsep}{0pt}
	\item \textit{GRPO$_{\text{full}}$.} Our full approach with all three edit-level classifiers ($c_{sim}$, $c_{flu}$, $c_{con}$).
	\item \textit{GRPO$_{\text{sim/flu/con}}$.} Ablations using only a single edit-level classifier.
	\item \textit{GRPO$_{\text{no\_sim/no\_flu/no\_con}}$.} Ablations excluding one edit-level classifier.
	\item \textit{GRPO$_{\text{app}}$.} Using argument-level appropriateness reward without edit-level classifiers.
\end{itemize}

\subsection{Training Setup}
\label{sec:training-setup}

\begin{table*}[t]
\centering
\small
\renewcommand{\arraystretch}{1}
\setlength{\tabcolsep}{3pt}
\begin{tabularx}{\textwidth}{Xcrrrrrcccccccccccc}
\toprule
\textbf{Approach} & & \multicolumn{5}{c}{\textbf{Edit-Level Metrics}} & & \multicolumn{11}{c@{}}{\textbf{Argument-Level Metrics}} \\
\cmidrule(l@{1pt}r@{1pt}){3-7} \cmidrule(l@{1pt}r@{0pt}){9-19}
& & \bf Sim & \bf Flu & \bf Con & \bf HL & \bf \#HL & & \bf BS & \bf BS$_{\text{HL}}$ & & \bf PPL$\downarrow$ & \bf PPL$_{\text{HL}}$$\downarrow$ & & \bf App & \bf App$_{\text{HL}}$ & & \bf All & \bf All$_{\text{HL}}$ \\
\midrule
\multicolumn{19}{@{}l}{\textbf{Baseline} \citep{ziegenbein-etal-2024-llm}\protect\footnotemark
} \\
Alpaca  & & 0.647 & 0.594 & \underline{0.949} & 0.347 & \underline{239} & & 0.619 & 0.941 & & 35.17 & 82.82 & & 0.329 & 0.267 & & 0.180 & 0.145 \\
+ PPO$_{app}$ & & 0.335 & \underline{0.624} & 0.899 & 0.114 & 93 & & 0.191 & 0.947 & & \underline{\textbf{18.34}} & 83.31 & & \underline{0.720} & 0.271 & & 0.196 & 0.146 \\
+ PPO$_{app>sim}$ & & 0.382 & 0.576 & 0.878 & 0.156 & 99 & & 0.298 & 0.945 & & 24.65 & 83.24 & & 0.547 & 0.284 & & 0.188 & 0.148 \\
+ PPO$_{app=sim}$ & & 0.570 & 0.492 & 0.913 & 0.233 & 142 & & 0.436 & 0.942 & & 27.29 & 79.60 & & 0.551 & \underline{0.293} & & \underline{0.206} & \underline{0.151} \\
+ PPO$_{app<sim}$ & & \underline{0.782} & 0.613 & 0.931 & \underline{0.475} & 186 & & \underline{\textbf{0.829}} & \underline{\textbf{0.964}} & & 45.59 & \underline{78.68} & & 0.289 & 0.276 & & 0.174 & 0.150 \\
\addlinespace
\multicolumn{19}{@{}l}{\bf Our Approach} \\
LLaMA & & 0.823 & 0.659 & 0.931 & 0.543 & 960 & & 0.707 & \underline{0.864} & & 48.67 & 56.22 & & 0.404 & 0.298 & & 0.180 & 0.166 \\
+ GRPO$_{app}$ & & 0.662 & 0.489 & 0.897 & 0.316 & 449 & & 0.326 & 0.854 & & \underline{20.03} & 65.42 & & 0.898 & 0.324 & & 0.242 & 0.149 \\
+ GRPO$_{con}$ & & 0.729 & 0.446 & \underline{\textbf{0.975}} & 0.281 & 649 & & 0.292 & 0.826 & & 31.16 & 63.83 & & \underline{\textbf{0.902}} & 0.342 & & 0.204 & 0.164 \\
+ GRPO$_{flu}$ & & 0.610 & \underline{\textbf{0.872}} & 0.757 & 0.392 & 501 & & 0.465 & 0.814 & & 22.69 & 72.23 & & 0.787 & \underline{\textbf{0.373}} & & \underline{\textbf{0.253}} & 0.161 \\
+ GRPO$_{sim}$ & & 0.951 & 0.366 & 0.967 & 0.346 & 998 & & 0.376 & 0.828 & & 27.01 & 56.23 & & 0.773 & 0.364 & & 0.221 & 0.175 \\
+ GRPO$_{no\_con}$ & & 0.926 & 0.838 & 0.893 & \underline{\textbf{0.709}} & 996 & & \underline{0.745} & 0.836 & & 33.65 & 54.83 & & 0.431 & 0.347 & & 0.212 & 0.174 \\
+ GRPO$_{no\_flu}$ & & \underline{\textbf{0.953}} & 0.348 & \underline{\textbf{0.975}} & 0.333 & 1112 & & 0.406 & 0.825 & & 36.23 & 61.84 & & 0.760 & 0.351 & & 0.204 & 0.167 \\
+ GRPO$_{no\_sim}$ & & 0.700 & 0.774 & 0.942 & 0.511 & 791 & & 0.545 & 0.823 & & 30.64 & 55.75 & & 0.667 & 0.369 & & 0.228 & 0.176 \\
+ \textbf{GRPO$_{full}$} & & 0.915 & 0.757 & 0.939 & 0.669 & \underline{\textbf{1221}} & & 0.742 & 0.842 & & 38.50 & \underline{\textbf{50.98}} & & 0.422 & 0.364 & & 0.201 & \underline{\textbf{0.182}} \\
\midrule
Gemini 2.5 & & 0.795 & 0.656 & 0.971 & 0.499 & 743 & & 0.675 & 0.873 & & 56.29 & 66.09 & & 0.511 & 0.347 & & 0.183 & 0.166 \\
GPT-5 & & 0.935 & 0.675 & 0.975 & 0.618 & 1683 & & 0.630 & 0.807 & & 43.12 & 54.34 & & 0.618 & 0.396 & & 0.208 & 0.181 \\
\bottomrule
\end{tabularx}
\caption{Main automatic evaluation results, comparing our approach \emph{GRPO$_{full}$} and various ablations to the baseline of \citet{ziegenbein-etal-2024-llm} on the edit level (left) and on the argument level (right). Underline indicates best in group, bold best overall. For comparison, we show the results of two prompted closed LLMs in the bottom lines.\protect\footnotemark}
\label{tab:ablation}
\end{table*}

We fine-tune our policy $\pi_\theta$ (Llama-3.1-8B-Instruct \citep{grattafiori-etal-2024-llama}) using GRPO \citep{shao-etal-2024-deepseek} with LoRA \citep{hu-etal-2022-lora}. The policy generates JSON output containing edit suggestions. We prompt the model with category explanations and an example (see Appendix~\ref{sec:appendix-prompt}). Full training details are in Appendix~\ref{sec:appendix-training-setup}.

\subsection{Evaluation Setup}
\label{sec:evaluation-setup}

We evaluate our approach and baselines on the test set of 225 inappropriate arguments from the original corpus of \citet{ziegenbein-etal-2023-modeling} using the following metrics. In a follow-up experiment, we also evaluate iterative revisions by our approach. 

\paragraph{Edit-Level Metrics}
For each argument, we generate edit suggestions using the trained model and compute the following edit-level metrics:
\begin{itemize}
\setlength{\itemsep}{-1pt}
	\item \textit{Sim.} Proportion of edits maintaining sufficient semantic similarity to the original.
	\item \textit{Flu.} Proportion of edits maintaining or improving fluency.
	\item \textit{Con.} Proportion of edits exhibiting conform editing patterns.
	\item \textit{HL.} Proportion of edits passing all three criteria (Sim, Flu, Con).
	\item \textit{\#HL.} Absolute number of human-like edits.
\end{itemize}

\paragraph{Argument-Level Metrics}
On the argument level, we use the following metrics, also computed on the human-like edits only (subscript \emph{HL}):
\begin{itemize}
\setlength{\itemsep}{0pt}
	\item \textit{BS / BS$_{\text{HL}}$.} BERTScore \citep{zhang-etal-2020-bertscore} between the original and edited argument.
	\item \textit{PPL / PPL$_{\text{HL}}$.} Fluency in terms of perplexity.
	\item \textit{App / App$_{\text{HL}}$.} The percentage of arguments whose appropriateness classification is flipped from inappropriate to appropriate.
	\item \textit{All / All$_{\text{HL}}$.} Geometric mean of \emph{BS}, $1/$\emph{PPL}, and \emph{App} as a single overall score.
\end{itemize}

\section{Results}
\label{sec:evaluation}

This section discusses the results of our experiments. We first detail the quantitative results across various edit-level and argument-level metrics, followed by a qualitative analysis of generated edits.

\subsection{Automatic Evaluation}
\label{sec:quantitative-results}

Table~\ref{tab:ablation} opposes our approach (GRPO$_{full}$) and several ablations against the PPO baselines. 

For the baselines, appropriateness improvement (\emph{App}) varies substantially (0.289--0.720), while for human-like edits only (\emph{App$_{\text{HL}}$}), it remains nearly constant (0.267--0.293). This indicates that, as the PPO models improve overall appropriateness, their edits become less human-like. Alpaca produces the highest number of human-like edits (239) among the baselines, and PPO$_{app<sim}$ the highest proportion (0.475). In contrast, PPO$_{app=sim}$ achieves a better geometric mean (\emph{All} = 0.206, \emph{All$_{\text{HL}}$} = 0.151). This demonstrates that simply maximizing the amount of human-like edits is insufficient; a balanced reward configuration is required. Ultima-tely, the results reveal that traditional rewriting approaches can optimize document-level appropriateness but fail to maintain edit-level human-likeness.

Our approach addresses this gap. GRPO$_{full}$ achieves drastically more human-like edits (\emph{\#HL} = 1221, \emph{HL} = 0.669) than all baselines, while simultaneously improving \emph{App$_{\text{HL}}$} (0.364). Its ablations suggest that all components contribute: removing semantic similarity (GRPO$_{no\_sim}$) improves fluency (from 0.757 to 0.774) but reduces human-likeness (\emph{HL} = 0.511); removing fluency (GRPO$_{no\_flu}$) optimizes semantic similarity (0.953) but severely impacts human-likeness (\emph{HL} = 0.333); and removing conformity (GRPO$_{no\_con}$) maximizes human-likeness (\emph{HL} = 0.709) but reduces conformity. Single-reward variants (GRPO$_{sim}$, GRPO$_{flu}$, GRPO$_{con}$) all perform worse than GRPO$_{full}$, confirming that our multi-objective formulation successfully balances edit-level quality with human-like editing patterns.
GRPO$_{no\_con}$) maximizes human-likeness (\emph{HL} = 0.709) but reduces conformity.

\addtocounter{footnote}{-2}
\stepcounter{footnote}\footnotetext{App values differ slightly from \citet{ziegenbein-etal-2024-llm} due to formatting requirements for diff computation, but all trends remain consistent.}
\stepcounter{footnote}\footnotetext{The closed LLMs use the same prompt as our approach.}

\paragraph{Iterative Revisions}


To study the effect of iteratively revising arguments with LLMs, we reapply our approach to its own output until the proportion of arguments classified as appropriate (\emph{App}) converges. Figure~\ref{revision-rounds} shows all metrics over the resulting 11 revision rounds on the test set.

\bsfigure{revision-rounds}{Iterative revision results across 11 rounds. (a) Edit-level metrics. (b) Argument-level metrics. $\Delta$ shows the difference between first and last rounds.}

At the edit level, pattern conformity (\emph{Con}) and semantic similarity (\emph{Sim}) remain nearly stable ($\Delta = -0.013$ / $-0.068$). However, fluency (\emph{Flu}) and, hence, human-likeness (\emph{HL}) decrease notably ($\Delta = -0.304$ / $-0.294$). This suggests that generating human-like edits becomes increasingly difficult in later rounds, primarily due to the progressive degradation of fluency.

In contrast to the stability observed at the edit level, argument-level similarity exhibits a noticeable decline ($\Delta=-0.243$ / $\Delta_{HL}=-0.284$). So, while individual edits remain locally similar to the input, their accumulation results in a meaning drift across the argument. Fluency also degrades ($\Delta=-0.119$), with a more pronounced drop when restricting revisions to the human-like subset ($\Delta_{HL}=-0.238$). However, as intended, the percentage of arguments classified as appropriate rises substantially across revisions ($\Delta=0.240$ / $\Delta_{HL}=0.263$). Ultimately, these results suggest that iterative application drives the text toward a generic state of appropriateness at the cost of preserving the argument's original intent and linguistic fluency. For all scores, see Appendix~\ref{app:interative-results}.

\subsection{Human Evaluation}
\label{sec:qualitative-results}

We conducted two human evaluation studies, each with three English native speakers, to complement our quantitative analysis.%
\footnote{The human evaluators were hired on \url{https://upwork.com} and compensated at \$15 per hour.}
Together, the two studies yield 4,200 individual annotations (2,400 in the first study and 1,800 in the second).

In the first study, the evaluators rated 200 edit suggestions from three approaches (PPO$_{app=sim}$, LLaMA, and GRPO$_{full}$) on the four edit-level metrics using a 5-point Likert scale, each rated by all three annotators (Krippendorff's $\alpha = 0.3317$). Table~\ref{tab:human-eval-quality} shows that GRPO$_{full}$ achieves the highest ratings across all dimensions (\emph{Sim} = 4.41, \emph{Flu} = 4.46, \emph{Con} = 4.24, \emph{HL} = 4.46), outperforming PPO$_{app=sim}$ and also slightly LLaMA, demonstrating that our approach maintains semantic similarity while improving fluency and pattern conformity.

In the second study, the evaluators performed pairwise comparisons in terms of overall quality for 100 arguments and their revised versions from four approaches (PPO$_{app=sim}$, LLaMA, GRPO$_{full}$ after one revision iteration, and GRPO$_{full}$ after 11 iterations). Table~\ref{tab:human-eval-ranking} presents rankings computed using the Bradley-Terry model \citep{bradley1952rank} with moderate inter-annotator agreement (Pearson's $r = .552$). The edit suggestions of GRPO$_{full}$ (11$^{th}$) are ranked best (\emph{p} = .903, \emph{Avg} = 1.34), being first in 82\% of comparisons, while PPO$_{app=sim}$ is last in 72\% (\emph{p} = .070). This underlines the effectiveness of our approach. Appendix~\ref{app:category-violations} gives examples of good and bad edit suggestions of the different approaches.

\begin{table}[t]
\centering
\small
\renewcommand{\arraystretch}{1.0}
\setlength{\tabcolsep}{6pt}
\begin{tabular}{lcccc}
\toprule
\textbf{Approach} 	& \textbf{Sim} 	& \textbf{Flu} 	& \textbf{Con} 	& \textbf{HL} \\
\midrule
PPO$_{app=sim}$ 	& 3.13 		& 3.48\phantom{$^{\dagger}$} 		& 3.75 		& 3.97 \\
LLaMA 			& 4.29 		& 4.38\phantom{$^{\dagger}$} 		& 4.21 		& 4.42 \\
+ \textbf{GRPO$_{full}$} (approach)
				& \textbf{4.41} 	& \textbf{4.46}$^{\dagger}$ 	& \textbf{4.24} 	& \textbf{4.46} \\
\bottomrule
\end{tabular}
\caption{Human evaluation of edit suggestions in terms of similarity, fluency, conformity, and human-likeness. Our approach \textit{GRPO$_{full}$} is best (bold) in all dimensions. $^{\dagger}$Significantly better than LLaMA ($p < .05$).}
\label{tab:human-eval-quality}
\end{table}


\begin{table}[t]
\centering
\small
\renewcommand{\arraystretch}{1.0}
\setlength{\tabcolsep}{3pt}
\begin{tabular}{lrrrrrr}
\toprule
\textbf{Model} & \textbf{1st} & \textbf{2nd} & \textbf{3rd} & \textbf{4th} & \textbf{Avg}$\downarrow$ & \textbf{p}$\uparrow$ \\
\midrule
PPO$_{app=sim}$ & 1{\scriptsize\%} & 6{\scriptsize\%} & 21{\scriptsize\%} & \textbf{72{\scriptsize\%}} & 3.64 & .070 \\
LLaMA & 8{\scriptsize\%} & 33{\scriptsize\%} & \textbf{45{\scriptsize\%}} & 14{\scriptsize\%} & 2.65 & .405 \\
+ \textbf{GRPO$_{full}$} (1$^{st}$) & 9{\scriptsize\%} & \textbf{55{\scriptsize\%}} & 26{\scriptsize\%} & 10{\scriptsize\%} & 2.37 & .515 \\
+ \textbf{GRPO$_{full}$} (11$^{th}$) & \textbf{82{\scriptsize\%}} & 6{\scriptsize\%} & 8{\scriptsize\%} & 4{\scriptsize\%} & \textbf{1.34} & \textbf{.903} \\
\bottomrule
\end{tabular}
\caption{Human pairwise comparison rankings of revised arguments in terms of overall quality. \emph{p} is the Bradley-Terry model merit. Lowest value bold for \emph{Avg}, otherwise highest value in each column bold.}
\label{tab:human-eval-ranking}
\end{table}

\section{Discussion}
\label{sec:discussion}

We discuss two aspects of our approach: the fine-grained behavior of the reward classifiers across inappropriateness dimensions, and the trade-offs of our sentence-level edit formulation.

\subsection{Reward Classifiers by Category}

Table~\ref{tab:reward-by-category} shows reward classifier pass rates per inappropriateness dimension \citep{ziegenbein-etal-2023-modeling}. Fluency is the primary bottleneck across all models, with edits targeting \emph{Toxic Emotions} (deceptive or excessive emotional appeals) and \emph{Missing Commitment} (lack of seriousness or openness to discussion) requiring substantial linguistic changes, resulting in lower human-likeness rates for GRPO$_{full}$ (HL: 67.8\% and 72.5\%) than dimensions such as \emph{Missing Intelligibility} (unclear meaning or reasoning) or \emph{Other Reasons} (orthographic and other issues; HL: 76.2\% and 76.4\%). Importantly, GRPO$_{full}$ maintains stable similarity rates ($\approx$95\%) across all dimensions, in contrast to LLaMA, which shows more variation (87.5\%--93.9\%), and PPO$_{app=sim}$, which drops steeply in similarity for \emph{Missing Intelligibility} (36.4\%). This suggests that the edit-level reward structure successfully anchors the policy's edits to the original text, even for the most linguistically challenging dimensions. PPO$_{app=sim}$ also produces substantially fewer edit suggestions overall (e.g., 199 vs.\ 463 for \emph{Toxic Emotions}), reflecting its tendency toward fewer but larger-scope rewrites. Conformity remains consistently high ($\geq$93\%) across all models, confirming it as a dimension-agnostic signal.

\begin{table}[t]
\centering
\small
\renewcommand{\arraystretch}{1.1}
\setlength{\tabcolsep}{4pt}
\resizebox{\columnwidth}{!}{\begin{tabular}{llccccc}
\toprule
\textbf{Dim} & \textbf{Approach} & \textbf{\#} & \textbf{Sim} & \textbf{Flu} & \textbf{Con} & \textbf{HL} \\
\midrule
\multirow{3}{*}{TE}
  & GRPO$_{full}$   & 463 & \underline{95.2\%} & \underline{73.4\%} & 98.5\%             & \underline{67.8\%} \\
  & LLaMA           & 435 & 88.5\%             & 64.8\%             & \underline{99.1\%} & 55.6\%             \\
  & PPO$_{app=sim}$ & 199 & 48.2\%             & 50.8\%             & 95.5\%             & 21.6\%             \\
\addlinespace
\multirow{3}{*}{MC}
  & GRPO$_{full}$   & 218 & \textbf{\underline{96.8\%}} & \underline{75.2\%} & \textbf{\underline{99.5\%}} & \underline{72.5\%} \\
  & LLaMA           & 198 & 93.9\%                      & 66.7\%             & 99.0\%                      & 61.6\%             \\
  & PPO$_{app=sim}$ &  87 & 58.6\%                      & 50.6\%             & 95.4\%                      & 27.6\%             \\
\addlinespace
\multirow{3}{*}{MI}
  & GRPO$_{full}$   & 311 & \underline{94.9\%} & \textbf{\underline{84.2\%}} & 95.8\%             & \underline{76.2\%} \\
  & LLaMA           & 257 & 87.5\%             & 70.4\%                      & \underline{98.4\%} & 61.9\%             \\
  & PPO$_{app=sim}$ & 140 & 36.4\%             & 57.1\%                      & 93.6\%             & 12.9\%             \\
\addlinespace
\multirow{3}{*}{OR}
  & GRPO$_{full}$   & 382 & \underline{94.8\%} & \underline{80.9\%} & 99.0\%             & \textbf{\underline{76.4\%}} \\
  & LLaMA           & 353 & 91.8\%             & 75.6\%             & \underline{99.4\%} & 68.8\%             \\
  & PPO$_{app=sim}$ & 117 & 47.0\%             & 53.8\%             & 98.3\%             & 26.5\%             \\
\bottomrule
\end{tabular}}
\caption{Reward classifier pass rates per inappropriateness dimension (TE: Toxic Emotions, MC: Missing Commitment, MI: Missing Intelligibility, OR: Other Reasons). \emph{\#} is the number of edit suggestions per dimension. Bold marks the overall highest value per metric; underline marks the highest per dimension.}
\label{tab:reward-by-category}
\end{table}

\subsection{Sentence-Level vs.\ Discourse-Level Edits}

A natural question raised by our sentence-level formulation is whether the approach can accommodate argumentation flaws that require structural reorganization across sentences, for example by reordering claims or redistributing content between paragraphs. Two aspects of our design are relevant here. First, from a \emph{technical} standpoint, our framework is not strictly limited to intra-sentence changes: since the entire argument is processed in a single forward pass, the policy can generate edits that span sentence boundaries (e.g., deleting content in one sentence and inserting related content in another). Second, from a \emph{design} standpoint, the sentence-level decomposition is a deliberate choice rather than a constraint: knowing which sentence an edit operates in makes it unambiguous which span is targeted, even when the same word or phrase occurs multiple times across the argument.

The key trade-off is between precision and structural expressivity. Sentence-level edits excel at targeted, atomic corrections most useful in interactive writing support, but may not fully capture flaws that inherently require redistributing or reordering argumentative content. In practice, our manual inspection did not surface notable cases of this limitation, and the human evaluation confirms that GRPO$_{full}$ edit suggestions are preferred both at the edit and argument level. Addressing more complex structural reorganization remains a direction for future work.

\section{Conclusion}
\label{sec:conclusion}

Improving human-written text has become a standard use case of LLMs, yet LLMs follow different editing strategies. To teach LLMs huma-like editing,
we have presented the first reinforcement learning approach to generating edit suggestions for inappropriate arguments that explicitly optimizes for edit-level quality. By rewarding the semantic similarity, fluency, and pattern conformity of edits, our approach generates substantially more human-like edit suggestions than baseline approaches while still achieving strong argument-level appropriateness. Via multi-round iterative editing, it can further close in on the appropriateness improvement of the best non-human-like rewriting baselines.

Our ablation studies confirm that all three edit-level quality dimensions contribute to human-like editing, while our full approach achieves the best balance of edit quality and appropriateness improvement.
The results demonstrate that optimizing for edit-level quality is crucial for generating self-contained, selectively applicable edit suggestions, which is a key requirement for LLM-assisted text optimization. 
Thereby, our work establishes a foundation of automatic systems that provide users with actionable, human-like feedback for improving argumentative writing.

\section{Limitations}
\label{sec:limitations}

While our approach successfully teaches LLMs to generate human-like edit suggestions for inappropriate arguments, some limitations remain.

First, our definition of appropriateness and the corresponding reward model rely on existing datasets and annotations. Appropriateness is inherently subjective and culture-dependent. What is considered appropriate in one context or culture might differ in another. Our model, trained on specific data, may not generalize well to all cultural contexts or definitions of appropriateness.

Second, our approach focuses on sentence-level edits. While we ensure fluency and coherence within sentences, and aim for argument-level appropriateness, we do not explicitly model discourse-level structure or inter-sentence dependencies beyond what the underlying LLM captures. Complex argumentation flaws that require restructuring the entire argument flow might not be fully addressed by sentence-level edits alone. Our manual inspection of sample output does not suggest notable issues in this regard, though.

Third, the reliance on proxy metrics for the reward function (semantic similarity, fluency, pattern conformity) introduces potential gaps. For instance, high semantic similarity is desirable to preserve meaning, but correcting highly inappropriate content might inherently require significant semantic changes. Balancing these trade-offs remains a challenge that may not be feasible perfectly in all cases.

Finally, the transferability of our approach to other languages with different syntactic structures or editing patterns remains to be investigated, as our experiments focus on English arguments.

\section{Ethical Considerations}
\label{sec:ethics}

The development of automated methods and tools for editing argumentation raises important ethical concerns that must be addressed to ensure responsible deployment.

A primary concern is the potential misuse of such systems for censorship or ``tone policing.'' By automating the process of making arguments ``appropriate,'' the nuances of passionate or culturally-specific forms of expression could be sanitized or suppressed. If deployed without careful oversight, such tools could disproportionately affect marginalized groups whose discourse styles may differ from the ``standard'' norms encoded in the training data. As in other reward-based systems, the idea of our approach could be inverted: the reward signals could be reversed to train models that intentionally generate harmful edits, making arguments deliberately inappropriate rather than appropriate. However, since both appropriate and inappropriate arguments are crucial to developing the rewriting approach, we see no way around this but to strongly emphasize not using the approach for this purpose.

Furthermore, the notion of \emph{appropriateness} is not universal; it is shaped by social, cultural, and political contexts. Our models are trained on datasets that reflect specific definitions of appropriateness, which inevitably contain biases. Consequently, the model's edit suggestions may enforce a specific worldview or linguistic standard. Users should be made aware that the suggestions reflect a specific model of appropriateness that may not be applicable or desirable in all contexts.

Given these risks, we emphasize the importance of a human-in-the-loop \emph{deployment} strategy. We stress that this is an ethical recommendation for responsible real-world use, not a technical requirement: the GRPO$_{full}$ framework is fully automated and generates edit suggestions autonomously without any human intervention during inference. In deployment, the system is designed to generate suggestions that a human author can accept, reject, or modify. It should not be used to automatically rewrite content without human review. The goal is to empower users to refine their arguments, not to replace their voice.

\section*{Acknowledgments}
\label{sec:acknowledgements}

The computational experiments in this paper have been partially supported by the 
Federal Ministry of Research, Technology, and Space (BMFTR),
Germany, as part of the AI service center KISSKI (grant number 01IS22093C).

\bibliography{acl26-edit-lit}

\appendix
\appendix

\clearpage
\onecolumn

\section{Prompt Template}
\label{sec:appendix-prompt}

\vspace{0.5em}
\noindent\textbf{Task:} Analyze the following argument by breaking it down into individual sentences. For each sentence, identify all inappropriate parts and edit it to make it appropriate while preserving the author's core point.

\vspace{0.5em}
\noindent The output must be a single JSON object with a single key ``sentence\_edits''. The value of this key should be a list of objects. Each object in the list must correspond to a sentence from the original argument and contain three keys in this specific order:

\begin{itemize}[leftmargin=1.5em, itemsep=2pt, topsep=2pt]
    \item \textbf{sentence\_id}: The sentence number (e.g., 1, 2, 3) corresponding to the input sentence.
    \item \textbf{rewritten\_sentence}: The full, clean, and fluent version of the rewritten sentence.
    \item \textbf{edits}: A list of JSON objects, where each object represents a single correction and contains three keys: ``inappropriate\_part'', ``rewritten\_part'', and ``reason''. The ``reason'' must be one of the allowed reason values.
\end{itemize}

\vspace{0.5em}
\noindent\textbf{Definitions for inappropriateness reasons:}

\begin{itemize}[leftmargin=1.5em, itemsep=2pt, topsep=2pt]
    \item \textbf{Toxic Emotions}: Emotions appealed to are deceptive or so intense that they discourage critical evaluation.
    \item \textbf{Missing Commitment}: The issue is not taken seriously or there is no openness to others' arguments.
    \item \textbf{Missing Intelligibility}: Meaning is unclear/irrelevant or reasoning is not understandable.
    \item \textbf{Other Reasons}: Severe orthographic errors or other issues not covered above.
\end{itemize}

\vspace{0.5em}
\noindent\textbf{Allowed Reason values:} Toxic Emotions, Missing Commitment, Missing Intelligibility, Other Reasons

\vspace{0.5em}
\noindent\textbf{Example:}

\noindent\textit{Issue:} Pro choice vs pro life

\noindent\textit{Input Sentences:}
\begin{itemize}[leftmargin=1.5em, itemsep=2pt, topsep=2pt]
    \item Sentence 1: for everyone who is talking about RAPE, let me ask you one thing!!!!
    \item Sentence 2: if you got in a huge fight with someone and ended up breaking your hand or arm ... would you cut it off just because it would REMIND you of that expirience???
    \item Sentence 3: if your actualy SANE you would say no and if you say yes you need to see a Physiatrist!!!!
\end{itemize}

\noindent\textit{JSON Output:}

\begin{small}
\begin{verbatim}
{
    "sentence_edits": [
        {
            "sentence_id": 1,
            "rewritten_sentence": "For those discussing rape,
                                   consider this:",
            "edits": [
                {
                    "inappropriate_part": "for everyone who is
                                           talking about",
                    "rewritten_part": "For those discussing",
                    "reason": "Missing Intelligibility"
                },
                {
                    "inappropriate_part": "RAPE",
                    "rewritten_part": "rape",
                    "reason": "Toxic Emotions"
                },
                {
                    "inappropriate_part": ", let me ask you one
                                           thing!!!!",
                    "rewritten_part": ", consider this:",
                    "reason": "Toxic Emotions"
                }
            ]
        },
        ... (additional sentences omitted for space)
    ]
}

\end{verbatim}
\end{small}

\vspace{0.5em}
\noindent Now complete the task for the following:

\vspace{0.3em}
\noindent\textit{Issue:} \{issue\}

\noindent\textit{Input Sentences:} \{sentences\}

\noindent\textit{JSON Output:}


\section{Gemini Fluency Augmentation Prompt}
\label{sec:appendix-fluency-prompt}

\vspace{0.5em}
\noindent\textbf{System Prompt:}

\vspace{0.3em}
\noindent You are a meticulous language editor. Your task is to evaluate a suggested text modification for its impact on sentence fluency.

\vspace{0.5em}
\noindent\textbf{Objective:} Given an original sentence, a specific part to be replaced, and the replacement text, you must determine if the resulting new sentence is \textbf{at least as fluent} as the original.

\vspace{0.5em}
\noindent\textbf{Definition of Fluency:} A sentence is considered ``fluent'' if it is grammatically correct, natural-sounding, easy to read, and clear in its meaning. An edit is acceptable if it maintains or improves fluency. An edit is unacceptable if it harms fluency in any way (e.g., makes it ungrammatical, awkward, or less clear).

\vspace{0.5em}
\noindent\textbf{Instructions}
\begin{enumerate}[leftmargin=1.5em, itemsep=2pt, topsep=2pt]
    \item \textbf{Reconstruct the Sentence:} Mentally replace the \texttt{\{inappropriate\_part\}} with the \texttt{\{rewritten\_part\}} in the \texttt{\{original\_sentence\}} to create the \texttt{New Sentence}.
    \item \textbf{Compare:} Carefully compare the \texttt{Original Sentence} and the \texttt{New Sentence}.
    \item \textbf{Evaluate:} Judge whether the \texttt{New Sentence} is at least as fluent as the \texttt{Original Sentence}.
    \item \textbf{Respond:} Provide your answer in a JSON object with two keys:
    \begin{itemize}[leftmargin=1.5em, itemsep=2pt, topsep=2pt]
        \item \texttt{"is\_fluent"}: A boolean (\texttt{true} or \texttt{false}).
        \item \texttt{"reason"}: A brief, one-sentence explanation for your decision.
    \end{itemize}
\end{enumerate}

\vspace{0.5em}
\noindent\textbf{Task}

\noindent Now, evaluate the following input:

\begin{itemize}[leftmargin=1.5em, itemsep=2pt, topsep=2pt]
    \item \textbf{Original Sentence:} \texttt{\{original\_sentence\}}
    \item \textbf{Substring to Replace:} \texttt{\{inappropriate\_part\}}
    \item \textbf{Replacement Substring:} \texttt{\{rewritten\_part\}}
\end{itemize}

\vspace{0.3em}
\noindent\textbf{Your JSON Output:}

\clearpage

\twocolumn

\section{Fluency Classifier}
\label{sec:appendix-fluency}

\subsection{Training Setup}

We train a binary classifier based on ModernBERT to distinguish between fluent and non-fluent edits. The classifier takes as input a sentence pair (original and edited) and outputs a binary decision. Training data is described in Section~\ref{sec:data}. We found that using only the GEC data from IteraTeR was insufficient to achieve reliable results, as the classifier struggled to generalize to the types of edits encountered during GRPO training, motivating the augmentation with Gemini-collected instances (see Section~\ref{sec:appendix-fluency-prompt} for the prompt). We use a 70/10/20 split for training, validation, and test data. The classifier is trained with a focus on high precision, as false positives (accepting non-fluent edits) are more harmful to GRPO training than false negatives (rejecting fluent edits). We train for 3 epochs with batch size 16 and learning rate $2 \times 10^{-5}$, using the AdamW optimizer with cosine learning rate schedule and warmup ratio of 0.1. We use BF16 precision and binary cross-entropy loss with class weighting to optimize for precision while maintaining reasonable recall.

\subsection{Evaluation Setup}

Detecting fluency degradation caused by edits is a novel task that differs from traditional grammatical error detection or absolute fluency assessment. Unlike these tasks, edits are not expected to fix all fluency issues in a sentence, nor do they necessarily need to improve fluency. It is sufficient if an edit fits into the sentence it is applied to without introducing new grammatical errors. This requires evaluating the relative change in fluency rather than the absolute fluency of the result. To validate our approach, we develop several baseline methods adapted from related tasks and compare them against our ModernBERT-based classifier. Table~\ref{tab:fluency-comparison-appendix} presents results on a combined test set of 3,665 examples.

\begin{table}[h]
\centering
\small
\renewcommand{\arraystretch}{1.1}
\setlength{\tabcolsep}{7pt}
\begin{tabular}{lcccc}
\toprule
\textbf{Approach} & \textbf{Acc.} & \textbf{Prec.} & \textbf{Rec.} & \textbf{F$_1$} \\
\midrule
Random & .429 & .444 & .267 & .333 \\
Always fluent & .536 & .536 & \textbf{1.000} & .698 \\
Always non-fluent & .464 & -- & .000 & .000 \\
\addlinespace
LanguageTool & .535 & .442 & .948 & .599 \\
RoBERTa + Flan-T5 & .698 & .563 & .814 & .665 \\
Gemini 2.5 Flash & .874 & .795 & .906 & \textbf{.847} \\
\addlinespace
ModernBERT (ours) & \textbf{.890} & \textbf{.880} & .796 & .835 \\
\bottomrule
\end{tabular}
\caption{Performance comparison of fluency detection approaches in terms of accuracy (Acc.), precision (Prec.), recall (Rec.), and F1 score. Bold indicates best performance.}
\label{tab:fluency-comparison-appendix}
\end{table}

\subsubsection{Approaches}

Apart from trivial baselines (random, always fluent, and always non-fluent), we develop three baseline methods adapted from related tasks to establish the difficulty of this novel problem:

\begin{itemize}[leftmargin=1.5em, itemsep=0pt, topsep=2pt]
    \item \textit{LanguageTool (rule-based):} We adapt the rule-based LanguageTool\footnote{\url{https://languagetool.org/}} grammar checker by comparing error counts before and after editing, classifying an edit as fluent if the error count does not increase.
    \item \textit{RoBERTa + Flan-T5 (cascade):} We develop a cascaded approach using RoBERTa-based grammar error detection \citep{morris2020textattack} and Flan-T5 based grammar correction\footnote{\url{https://huggingface.co/pszemraj/flan-t5-large-grammar-synthesis}} on the edited sentence and check if corrections overlap with the edit region.
    \item \textit{Gemini 2.5 Flash (LLM):} We prompt Gemini 2.5 Flash \citep{comanici2025gemini25pushingfrontier} to evaluate whether the edited sentence is at least as fluent as the original.
    \item \textit{ModernBERT (trained classifier):} We train a ModernBERT-based \citep{warner-etal-2025-smarter} classifier on paired before/after sentences to detect fluency degradation, as described in the training setup.
\end{itemize}

\subsubsection{Evaluation}

Table~\ref{tab:fluency-comparison-appendix} presents the performance comparison across all approaches. For GRPO training, high precision is critical as false positives (accepting non-fluent edits) are more harmful than false negatives (rejecting fluent edits). The rule-based approach LanguageTool achieves high \emph{recall} (\emph{0.948}) but low \emph{precision} (\emph{0.442}), likely due to limited coverage—edits introducing subtle errors not covered by the rule set may be incorrectly accepted. The cascaded approach combining RoBERTa and Flan-T5 also focuses on recall (\emph{0.814}) rather than precision (\emph{0.563}) and may suffer from cascading errors and ambiguity in determining whether corrections overlap with the edit region. LLM-based prompting with Gemini 2.5 Flash achieves the best F1 score (\emph{0.847}) but exhibits lower \emph{precision} (\emph{0.795}) than the ModernBERT-based classifier, which best meets the requirements with \emph{0.880 precision} while maintaining the highest overall accuracy (\emph{0.890}) and second highest F1 score (\emph{0.835}).

\section{Pattern Conformity Classifier}
\label{sec:appendix-conformity}


The model is a transformer-based language model with 2 layers, 2 attention heads, embedding dimension 200, hidden dimension 200, and dropout 0.2, totaling 486,406 parameters. We use the IteraTeR dataset as described in Section~\ref{sec:data}. Training uses batch size 64, learning rate 0.001, maximum sequence length 500, and 5 epochs. The vocabulary size is 6 (five edit operations plus a pad token).

\section{GRPO Training Setup}
\label{sec:appendix-training-setup}

We fine-tune our policy $\pi_\theta$, instantiated as Llama-3.1-8B-Instruct, using GRPO with Low-Rank Adaptation (LoRA) for parameter-efficient training. The LoRA configuration uses rank 16, alpha value of 32, and dropout 0.1, applied to all linear layers of the LLM. We train on the inappropriate arguments of the appropriateness corpus extension dataset with 2 epochs, per-device batch size 2, and gradient accumulation steps 8 (effective batch size 16). For each instance in a batch, we explore eight different episodes. In total, we explore 568,592 episodes during training. The optimizer is 8-bit paged AdamW with learning rate $5 \times 10^{-6}$ and cosine scheduler. The KL penalty coefficient is $\beta = 0.001857$. We use BF16 mixed precision and DR-GRPO loss. For efficient inference during training, we employ vLLM in colocate mode. The reward function combines global appropriateness reward and dense local appropriateness reward with $\alpha = 0.5$ (see Section~\ref{sec:approach}). To train a single model we used 4 A100 GPUs and trained for 72 hours.

Our reward formulation introduces only one hyperparameter beyond those of the underlying LLM: $\alpha$. We set $\alpha = 0.5$ for all experiments without tuning. The reward design is inherently stable due to a unidirectional dependency: the edit-level rewards are computed independently for each edit, while the argument-level reward is conditioned on the human-like subset. In practice, $\alpha$ primarily affects GRPO convergence speed rather than final performance.

\section{Edit Suggestion Examples}
\label{app:category-violations}

\definecolor{editred}{rgb}{0.7, 0.0, 0.0}
\definecolor{editgreen}{rgb}{0, 0.43, 0.58}

This appendix presents edit suggestion examples illustrating quality differences across approaches. Tables~\ref{table-agreement-nat}--\ref{table-agreement-good} illustrates cases with lowest scores for pattern conformity, semantic similarity, fluency, and human-likeness, followed by highest scores across all dimensions. The xamples were sampled randomly for these scores, balancing across approaches; when insufficient examples existed for specific score ranges, substitutions from other models were made. All ratings are from the human study (Section~\ref{sec:experimental-setup}). Dotted underlines mark edit regions: \textcolor{editred}{red} for deletions, \textcolor{editgreen}{petrol} for insertions/replacements, black for unchanged tokens.

\paragraph{Pattern Conformity.} Low conformity scores (Table~\ref{table-agreement-nat}) indicate edit patterns that diverge from human editing conventions. Violations include extensive rewrites that add substantial new content (e.g., replacing ``ways to avoid the situation of'' with an entire sentence about adoption options), scattered multi-span changes across large text portions (e.g., modifying multiple non-adjacent tokens within a single edit), and fundamental sentence restructuring rather than targeted corrections.

\paragraph{Fluency.} Fluency violations (Table~\ref{table-agreement-flu}) manifest as grammatical errors introduced by the edit. Examples include creating duplicate words (``If it wa s wrong then''), breaking grammatical structure through unnecessary token additions (``Dosen't it says''), awkward phrasings, capitalization errors, and incomplete constructions that lack necessary complements (``some children may not understand the value of'' followed by unrelated text).

\paragraph{Similarity.} Severe similarity violations (Table~\ref{table-agreement-sim}) occur when edits completely replace utterances with semantically unrelated content, failing to preserve the author's original argument. Examples include substituting ``Sorry'' with an entirely different sentence about self-protection, replacing ``This is no different'' with an unrelated statement about government mandates, or fundamentally changing the subject and meaning of complex sentences (``You are better off...'' $\rightarrow$ ``It's better to...'').

\paragraph{Human-likeness.} High human-likeness scores (Table~\ref{table-agreement-good}) indicate edits that fulfill all quality criteria simultaneously (Pattern Conformity, Similarity, and Fluency). These examples demonstrate focused, minimal edits: simple capitalization corrections (``i'' $\rightarrow$ ``I''), removal of redundant phrases (``United States of America'' $\rightarrow$ ``United States''), or single-word substitutions that preserve meaning while improving appropriateness (``not mix'' $\rightarrow$ ``remain separate'').

\onecolumn
\clearpage
 \normalem  

\definecolor{editred}{rgb}{0.93, 0.11, 0.14}
\definecolor{editgreen}{rgb}{0, 0.43, 0.58}

\begingroup
\noindent
\small
\centering
\renewcommand{\arraystretch}{1.2}
\setlength{\tabcolsep}{3pt}
\begin{tabular}{p{0.12\textwidth}>{\raggedright\arraybackslash}p{0.845\textwidth}}
    \toprule
\textbf{Model} & \textbf{Example} \\
\midrule
\textbf{PPO$_{app=sim}$} & Having a lousy father \textcolor{editred}{\dotuline{you}} have a father that cares for you and loves you.\\
& $\hookrightarrow$ Having a lousy father \textcolor{editgreen}{\dotuline{\textcolor{editgreen}{\textbf{, }}\textcolor{black}{you}\textcolor{editgreen}{\textbf{ can still}}}} have a father that cares for you and loves you.\\
\midrule
\textbf{PPO$_{app=sim}$} & There are \textcolor{editred}{\dotuline{\textcolor{editred}{\textbf{ways to avoid the situation of}}}} abortion.\\
& $\hookrightarrow$ There are \textcolor{editgreen}{\dotuline{\textcolor{editgreen}{\textbf{thousands of people waiting to adopt a baby, so the woman has the option of giving the}}\newline\textcolor{editgreen}{\textbf{baby up for adoption instead of having an}}}} abortion.\\
\midrule
\textbf{LLaMA} & Ultimately, people don't help out people unless they get some sort of satisfaction out of it, be it whether they community service hours, \textcolor{editred}{\dotuline{\textcolor{black}{material rewards, fame, or }\textcolor{editred}{\textbf{just having }}\textcolor{black}{the satisfaction of knowing th}\textcolor{editred}{\textbf{at you hav}}\textcolor{black}{e importan}\textcolor{editred}{\textbf{ce}}}}.\\
& $\hookrightarrow$ Ultimately, people don't help out people unless they get some sort of satisfaction out of it, be it whether they community service hours, \textcolor{editgreen}{\dotuline{\textcolor{black}{material rewards, fame, or the satisfaction of knowing th}\textcolor{editgreen}{\textbf{ey'r}}\textcolor{black}{e importan}\textcolor{editgreen}{\textbf{t}}}}.\\
\midrule
\textbf{LLaMA} & \textcolor{editred}{\dotuline{\textcolor{black}{We're in Britain}\textcolor{editred}{\textbf{ here}} \textcolor{black}{where }\textcolor{editred}{\textbf{the }}\textcolor{black}{covering }\textcolor{editred}{\textbf{your}} \textcolor{black}{face is seen as suspicious}\textcolor{editred}{\textbf{,}} \textcolor{black}{these women }\textcolor{editred}{\textbf{relly }}\textcolor{black}{need to }\textcolor{editred}{\textbf{think about}} \textcolor{black}{the message the veil }\textcolor{editred}{\textbf{puts acros}}\textcolor{black}{s about them and their religion.}}}\\
& $\hookrightarrow$ \textcolor{editgreen}{\dotuline{\textcolor{black}{We're in Britain}\textcolor{editgreen}{\textbf{,}} \textcolor{black}{where covering }\textcolor{editgreen}{\textbf{one's}} \textcolor{black}{face is seen as suspicious}\textcolor{editgreen}{\textbf{;}} \textcolor{black}{these women need to }\textcolor{editgreen}{\textbf{consider}} \textcolor{black}{the message the veil }\textcolor{editgreen}{\textbf{send}}\textcolor{black}{s about them and their religion.}}}\\
\midrule
\textbf{+ GRPO$_{full}$} & Why, gay man here, can be fired for being gay at any job in the country, \textcolor{editred}{\dotuline{\textcolor{editred}{\textbf{can't have}} \textcolor{black}{a civil union with my partner of 9 years}}}, no domestic partnet rights, hell if it were up to some I would be better of on some deserted island with like kind........not free for me.......but still grateful to be an american and for those fighting for what rights i do have\\
& $\hookrightarrow$ Why, gay man here, can be fired for being gay at any job in the country, \textcolor{editgreen}{\dotuline{\textcolor{editgreen}{\textbf{denied}} \textcolor{black}{a civil union with my partner of 9 years}}}, no domestic partnet rights, hell if it were up to some I would be better of on some deserted island with like kind........not free for me.......but still grateful to be an american and for those fighting for what rights i do have\\
\bottomrule
\end{tabular}
\captionof{table}{Exemplary edit suggestions with \textbf{lowest} scores for \emph{pattern conformity}, generated by the respective model on the left.}
\label{table-agreement-nat}

\vspace{1em}

\noindent
\small
\centering
\renewcommand{\arraystretch}{1.2}
\setlength{\tabcolsep}{3pt}
\begin{tabular}{p{0.12\textwidth}>{\raggedright\arraybackslash}p{0.845\textwidth}}
    \toprule
\textbf{Model} & \textbf{Example} \\
\midrule
\textbf{PPO$_{app=sim}$} & \textcolor{editred}{\dotuline{\textcolor{editred}{\textbf{i}}\textcolor{black}{f it }\textcolor{editred}{\textbf{i}}\textcolor{black}{s wrong }\textcolor{editred}{\textbf{now}}}} it was wrong then.\\
& $\hookrightarrow$ \textcolor{editgreen}{\dotuline{\textcolor{editgreen}{\textbf{I}}\textcolor{black}{f it }\textcolor{editgreen}{\textbf{wa}}\textcolor{black}{s wrong }\textcolor{editgreen}{\textbf{then,}}}} it was wrong then.\\
\midrule
\textbf{LLaMA} & Dosen't \textcolor{editred}{\dotuline{\textcolor{black}{it say}}} in the declaration of independence that we have the right to life liberty and the purist of happiness\\
& $\hookrightarrow$ Dosen't \textcolor{editgreen}{\dotuline{\textcolor{black}{it say}\textcolor{editgreen}{\textbf{s}}}} in the declaration of independence that we have the right to life liberty and the purist of happiness\\
\midrule
\textbf{LLaMA} & \textcolor{editred}{\dotuline{\textcolor{editred}{\textbf{I understand a}}\textcolor{black}{s a union }\textcolor{editred}{\textbf{guy}}}} that our dues are paltry, and the teamsters and other unions supply a comparatively tiny amount of money to politicians as opposed to corporations.\\
& $\hookrightarrow$ \textcolor{editgreen}{\dotuline{\textcolor{editgreen}{\textbf{A}}\textcolor{black}{s a union }\textcolor{editgreen}{\textbf{member}}}} that our dues are paltry, and the teamsters and other unions supply a comparatively tiny amount of money to politicians as opposed to corporations.\\
\midrule
\textbf{+ GRPO$_{full}$} & If god loves everyone then y does he hate gays and the bible is just a book, \textcolor{editred}{\dotuline{\textcolor{editred}{\textbf{h}}\textcolor{black}{ave you ever wondered that the on}\textcolor{editred}{\textbf{e}}}} cardinal who translated it changed some things to make "god" look good\\
& $\hookrightarrow$ If god loves everyone then y does he hate gays and the bible is just a book, \textcolor{editgreen}{\dotuline{\textcolor{editgreen}{\textbf{H}}\textcolor{black}{ave you ever wondered that the }\textcolor{editgreen}{\textbf{pers}}\textcolor{black}{on}}} cardinal who translated it changed some things to make "god" look good\\
\midrule
\textbf{+ GRPO$_{full}$} & YES, because \textcolor{editred}{\dotuline{\textcolor{black}{some children }\textcolor{editred}{\textbf{don'}}\textcolor{black}{t understand }\textcolor{editred}{\textbf{anything}}}} expect physical education especially rich children of rich parents.\\
& $\hookrightarrow$ YES, because \textcolor{editgreen}{\dotuline{\textcolor{black}{some children }\textcolor{editgreen}{\textbf{may no}}\textcolor{black}{t understand }\textcolor{editgreen}{\textbf{the value of}}}} expect physical education especially rich children of rich parents.\\
\bottomrule
\end{tabular}
\captionof{table}{Exemplary edit suggestions with \textbf{lowest} scores for \emph{fluency}.}
\label{table-agreement-flu}
\endgroup

\vspace{3em}

\begingroup
\noindent
\small
\centering
\renewcommand{\arraystretch}{1.2}
\setlength{\tabcolsep}{3pt}
\begin{tabular}{p{0.12\textwidth}>{\raggedright\arraybackslash}p{0.845\textwidth}}
    \toprule
\textbf{Model} & \textbf{Example} \\
\midrule
\textbf{PPO$_{app=sim}$} & \textcolor{editred}{\dotuline{\textcolor{editred}{\textbf{Sorry}}}}\\
& $\hookrightarrow$ \textcolor{editgreen}{\dotuline{\textcolor{editgreen}{\textbf{Even when sometimes you feel you need to in order to protect yourself.}}}}\\
\midrule
\textbf{PPO$_{app=sim}$} & \textcolor{editred}{\dotuline{\textcolor{black}{Th}\textcolor{editred}{\textbf{is is no different}}}}.\\
& $\hookrightarrow$ \textcolor{editgreen}{\dotuline{\textcolor{black}{Th}\textcolor{editgreen}{\textbf{e government also mandates that women must be screened for certain conditions}}\newline\textcolor{editgreen}{\textbf{if they want to run for office}}}}.\\
\midrule
\textbf{PPO$_{app=sim}$} & \textcolor{editred}{\dotuline{\textcolor{editred}{\textbf{Like that person who just said TV is awesome with many exclamation marks and in capitals}}}}.\\
& $\hookrightarrow$ \textcolor{editgreen}{\dotuline{\textcolor{editgreen}{\textbf{I have read books that have changed my life in a way that watching a TV show cannot, but at the}}\newline\textcolor{editgreen}{\textbf{same time I have watched shows that have left a lasting impact on me}}}}.\\
\midrule
\textbf{LLaMA} & \textcolor{editred}{\dotuline{\textcolor{editred}{\textbf{You are}} \textcolor{black}{better }\textcolor{editred}{\textbf{off not drinking or smoking, exercising 5 times a week, eating a balanced}}\newline\textcolor{editred}{\textbf{diet, chewing 15 times before swallowing your food, looking both ways before crossing the street,}}\newline\textcolor{editred}{\textbf{and not running with scissors}}}}.\\
& $\hookrightarrow$ \textcolor{editgreen}{\dotuline{\textcolor{editgreen}{\textbf{It's}} \textcolor{black}{better }\textcolor{editgreen}{\textbf{to follow these habits for your health and well-being, rather than being forced}}\newline\textcolor{editgreen}{\textbf{to do so by law}}}}.\\
\midrule
\textbf{+ GRPO$_{full}$} & \textcolor{editred}{\dotuline{\textcolor{editred}{\textbf{Eh, }}\textcolor{black}{Atheism is a p}\textcolor{editred}{\textbf{retty cool guy}}}}.\\
& $\hookrightarrow$ \textcolor{editgreen}{\dotuline{\textcolor{black}{Atheism is a p}\textcolor{editgreen}{\textbf{hilosophical position that has its own merits}}}}.\\
\bottomrule
\end{tabular}
\captionof{table}{Exemplary edit suggestions with \textbf{lowest} scores for \emph{semantic similarity}.}
\label{table-agreement-sim}

\vspace{0.5em}

\noindent
\small
\centering
\renewcommand{\arraystretch}{1.2}
\setlength{\tabcolsep}{3pt}
\begin{tabular}{p{0.12\textwidth}>{\raggedright\arraybackslash}p{0.845\textwidth}}
    \toprule
\textbf{Model} & \textbf{Example} \\
\midrule
\textbf{PPO$_{app=sim}$} & \textcolor{editred}{\dotuline{\textcolor{editred}{\textbf{We like to scrimp if we can, so we can pocket the difference}}}}.\\
& $\hookrightarrow$ \textcolor{editgreen}{\dotuline{\textcolor{editgreen}{\textbf{If it was that useless, or even if Congress felt they could get away without it, they would}}\newline\textcolor{editgreen}{\textbf{immediately remove it}}}}.\\
\midrule
\textbf{PPO$_{app=sim}$} & \textcolor{editred}{\dotuline{\textcolor{editred}{\textbf{yes!}}}}\\
& $\hookrightarrow$ \textcolor{editgreen}{\dotuline{\textcolor{editgreen}{\textbf{Our children are generally not as physically active as they should be and tend to eat unhealthy}}\newline\textcolor{editgreen}{\textbf{diets, which is a concern for us.}}}}\\
\midrule
\textbf{LLaMA} & I thick that book are better than TV is \textcolor{editred}{\dotuline{\textcolor{editred}{\textbf{it is better}}}} i can put you in a whole norther wold\\
& $\hookrightarrow$ I thick that book are better than TV is \textcolor{editgreen}{\dotuline{\textcolor{editgreen}{\textbf{because it allows you to immerse yourself in a northern world}}}} i can put you in a whole norther wold\\
\midrule
\textbf{LLaMA} & I think if \textcolor{editred}{\dotuline{\textcolor{black}{the government wanted him dead}}}, they would have offed him WAY before when he was actually inventing things which could possibly be related to weapons.\\
& $\hookrightarrow$ I think if \textcolor{editgreen}{\dotuline{\textcolor{editgreen}{\textbf{If }}\textcolor{black}{the government wanted him dead}}}, they would have offed him WAY before when he was actually inventing things which could possibly be related to weapons.\\
\midrule
\textbf{+ GRPO$_{full}$} & You can be a kind, compassionate person and bad shit \textcolor{editred}{\dotuline{\textcolor{editred}{\textbf{will }}\textcolor{black}{still }\textcolor{editred}{\textbf{happen to you}}}}.\\
& $\hookrightarrow$ You can be a kind, compassionate person and bad shit \textcolor{editgreen}{\dotuline{\textcolor{black}{still }\textcolor{editgreen}{\textbf{experience}}}}.\\
\bottomrule
\end{tabular}
\captionof{table}{Exemplary edit suggestions with \textbf{lowest} scores for \emph{human-likeness}.}
\label{table-agreement-hl}

\vspace{0.5em}

\noindent
\small
\centering
\renewcommand{\arraystretch}{1.2}
\setlength{\tabcolsep}{3pt}
\begin{tabular}{p{0.12\textwidth}>{\raggedright\arraybackslash}p{0.845\textwidth}}
    \toprule
\textbf{Model} & \textbf{Example} \\
\midrule
\textbf{PPO$_{app=sim}$} & Marriage and the government should \textcolor{editred}{\dotuline{\textcolor{editred}{\textbf{not mix}}}}.\\
& $\hookrightarrow$ Marriage and the government should \textcolor{editgreen}{\dotuline{\textcolor{editgreen}{\textbf{remain separate}}}}.\\
\midrule
\textbf{LLaMA} & The purpose of this restriction is to cut down on the amount of dumb jokes, so we can keep \textcolor{editred}{\dotuline{\textcolor{editred}{\textbf{the }}\textcolor{black}{quality }\textcolor{editred}{\textbf{of }}\textcolor{black}{debate and discourse}\textcolor{editred}{\textbf{ as high as possible}}}}.\\
& $\hookrightarrow$ The purpose of this restriction is to cut down on the amount of dumb jokes, so we can keep \textcolor{editgreen}{\dotuline{\textcolor{editgreen}{\textbf{high-}}\textcolor{black}{quality debate and discourse}}}.\\
\midrule
\textbf{+ GRPO$_{full}$} & \textcolor{editred}{\dotuline{\textcolor{black}{I }\textcolor{editred}{\textbf{AM}} \textcolor{black}{an atheist}}}\\
& $\hookrightarrow$ \textcolor{editgreen}{\dotuline{\textcolor{black}{I }\textcolor{editgreen}{\textbf{am}} \textcolor{black}{an atheist}\textcolor{editgreen}{\textbf{.}}}}\\
\midrule
\textbf{LLaMA} & \textcolor{editred}{\dotuline{\textcolor{black}{In the United States}\textcolor{editred}{\textbf{ of America}}}}, pro-choice is the only constitutional way to go.\\
& $\hookrightarrow$ \textcolor{editgreen}{\dotuline{\textcolor{black}{In the United States}}}, pro-choice is the only constitutional way to go.\\
\midrule
\textbf{LLaMA} & \textcolor{editred}{\dotuline{\textcolor{editred}{\textbf{i}}}} remember the only country by whose fear alexander's army was afraid it was INDIA [..]\\
& $\hookrightarrow$ \textcolor{editgreen}{\dotuline{\textcolor{editgreen}{\textbf{I}}}} remember the only country by whose fear alexander's army was afraid it was INDIA [..]\\
\bottomrule
\end{tabular}
\captionof{table}{Exemplary edit suggestions with \textbf{highest} scores across all dimensions.}
\label{table-agreement-good}
\endgroup
\clearpage

\clearpage

\section{Annotation Interfaces}
\label{app:annotation-interfaces}

\begingroup
\noindent
\centering
\includegraphics[width=0.85\textwidth]{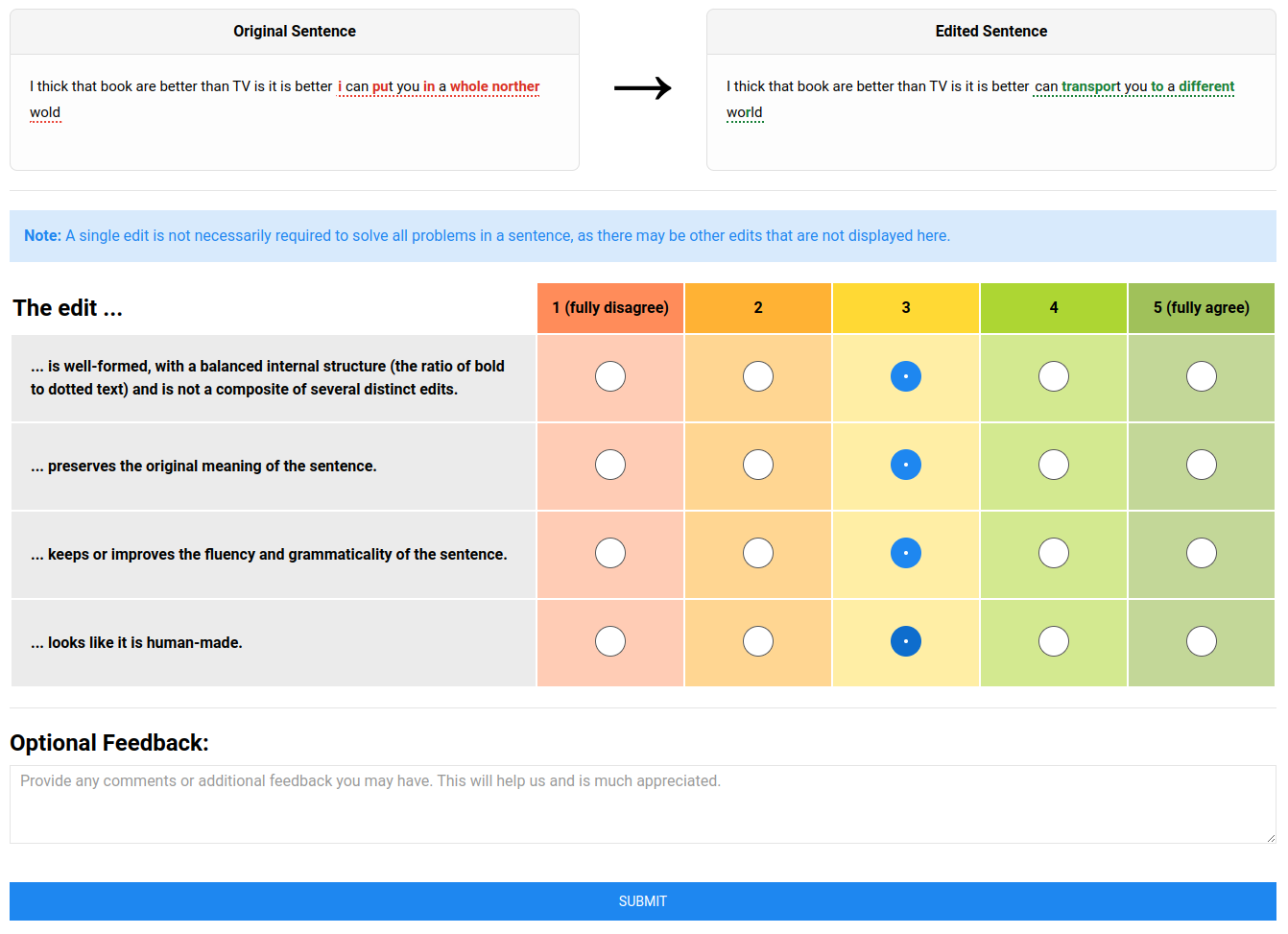}
\captionof{figure}{Annotation interface for edit suggestion quality ratings using a 5-point Likert scale (Pattern Conformity, Semantic Similarity, Fluency, and Human-Likeness). Screenshots of the full annotation guidelines can be found in the supplementary material.}
\label{fig:annotation-interface1}
\endgroup

\vspace{3.5em}

\begingroup
\noindent
\centering
\includegraphics[width=0.85\textwidth]{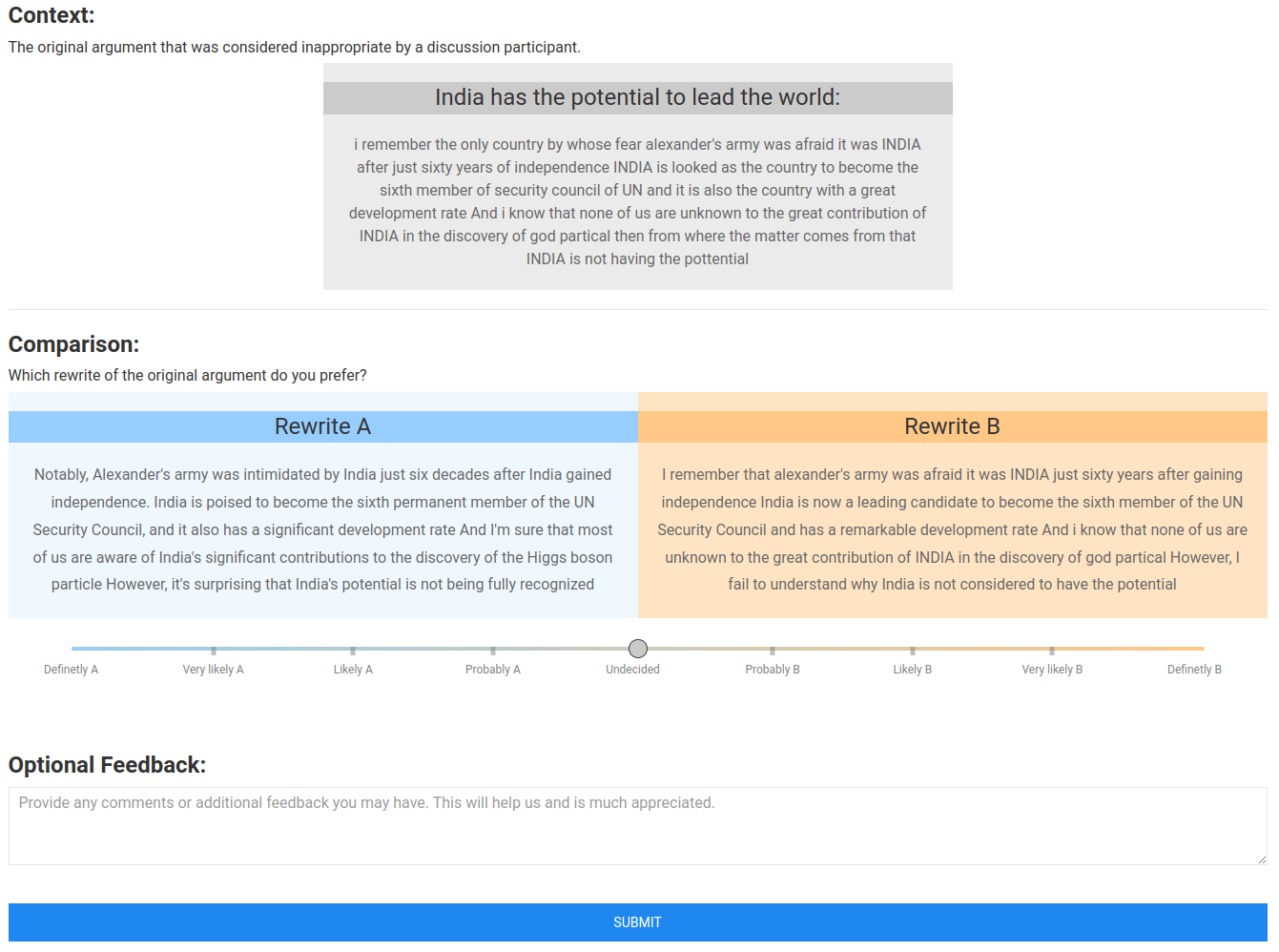}
\captionof{figure}{Annotation interface for pairwise comparison of arguments after applying edit suggestions. Screenshots of the full annotation guidelines can be found in the supplementary material.}
\label{fig:annotation-interface2}
\endgroup

\clearpage
\onecolumn

\section{Iterative Editing: Full Results}
\label{app:interative-results}

\begingroup
\noindent
\centering
\small
\renewcommand{\arraystretch}{1}
\setlength{\tabcolsep}{3pt}
\begin{tabular}{@{} l c rrrrr c c@{\hspace{3pt}}c c c@{\hspace{3pt}}c c c@{\hspace{3pt}}c c c@{\hspace{3pt}}c @{}}
\toprule
	\textbf{Revision} & & \multicolumn{5}{c}{\textbf{Edit-Level}} & & \multicolumn{11}{c@{}}{\textbf{Arg-Level}} \\
	\cmidrule(l@{1pt}r@{1pt}){3-7} \cmidrule(l@{1pt}r@{0pt}){9-19}
		& & Sim & Flu & Con & HL & \#HL & & BS & BS$_{\text{HL}}$ & & PPL & PPL$_{\text{HL}}$ & & App & App$_{\text{HL}}$ & & All & All$_{\text{HL}}$ \\
		\midrule
		1$^{st}$ (GRPO$_{full}$) & & \textbf{.915} & \textbf{.757} & .939 & \textbf{.669} & \textbf{1221} & & \textbf{.742} & \textbf{.842} & & 38.50 & 50.98 & & .422 & .364 & & .195 & .177 \\
		$\hookrightarrow$ 2$^{nd}$ & & .867 & .683 & .923 & .571 & 812 & & .688 & .764 & & 35.41 & 41.08 & & .467 & .422 & & .207 & .199 \\
		$\hookrightarrow$ 3$^{rd}$ & & .862 & .645 & .930 & .519 & 656 & & .650 & .719& & 31.47 & 36.66 & & .520 & .462 & & .218 & .208 \\
		$\hookrightarrow$ 4$^{th}$ & & .859 & .575 & .932 & .461 & 528 & & .617 & .684 & & 31.02 & 34.17 & & .564 & .471 & & .222 & .212 \\
		$\hookrightarrow$ 5$^{th}$ & & .853 & .592 & .914 & .467 & 519 & & .591 & .651 & & 29.26 & 32.59 & & .596 & .516 & & .229 & .219 \\
		$\hookrightarrow$ 6$^{th}$ & & .846 & .550 & .917 & .440 & 475 & & .568 & .623 & & 28.32 & 31.74 & & .613 & .542 & & .233 & .224 \\
		$\hookrightarrow$ 7$^{th}$ & & .843 & .503 & .915 & .404 & 401 & & .550 & .599 & & 27.46 & 30.52 & & .622 & .578 & & .235 & .229 \\
		$\hookrightarrow$ 8$^{th}$ & & .854 & .487 & \textbf{.946} & .376 & 370 & & .539 & .581 & & 28.20 & 29.95 & & .653 & .604 & & .237 & .233 \\
		$\hookrightarrow$ 9$^{th}$ & & .858 & .450 & .928 & .376 & 348 & & .524 & .567 & & 27.45 & 28.86 & & .627 & .604 & & .234 & .234 \\
		$\hookrightarrow$ 10$^{th}$ & & .835 & .473 & .934 & .351 & 334 & & .512 & .552 & & \textbf{26.29} & \textbf{26.91} & & .640 & .613 & & \textbf{.238} & .235 \\
		$\hookrightarrow$ 11$^{th}$ & & .847 & .453 & .926 & .375 & 325 & & .499 & .540 & & 26.62 & 27.24 & & \textbf{.662} & \textbf{.627} & & .236 & \textbf{.236} \\
	\bottomrule
\end{tabular}
\captionof{table}{Iterative editing results for GRPO$_{full}$ across 11 rounds on the edit level (left) and argument level (right). Only human-like edits are applied in each round until appropriateness converges. Bold indicates best values.}
\label{tab:iterative-editing}
\endgroup

\end{document}